\documentclass[sigconf]{acmart}

\AtBeginDocument{%
  \providecommand\BibTeX{{%
    \normalfont B\kern-0.5em{\scshape i\kern-0.25em b}\kern-0.8em\TeX}}}

\usepackage{subfigure}
\usepackage{color}
\usepackage{bbm}
\usepackage{multirow}

\copyrightyear{2021}
\acmYear{2021}
\setcopyright{acmcopyright}\acmConference[MM '21]{Proceedings of the 29th ACM
International Conference on Multimedia}{October 20--24, 2021}{Virtual Event, China}
\acmBooktitle{Proceedings of the 29th ACM International Conference on Multimedia
(MM '21), October 20--24, 2021, Virtual Event, China}
\acmPrice{15.00}
\acmDOI{10.1145/3474085.3475222}
\acmISBN{978-1-4503-8651-7/21/10}

\begin{document}
\fancyhead{}
\title{Two-stage Visual Cues Enhancement Network \\ for Referring Image Segmentation}

\author{Yang Jiao$^{1\ast}$ \quad  Zequn Jie$^{2}$ \quad Weixin Luo$^2$\quad Jingjing Chen$^{1\star}$}
\author{Yu-Gang Jiang$^1$ \quad Xiaolin Wei$^2$ \quad Lin Ma$^{2\star}$}
\affiliation{%
 \institution{$^1$Shanghai Key Lab of Intelligent Information Processing, School of Computer Science, Fudan University \hspace{0.3em} }
 \institution{$^2$Meituan \hspace{0.3em} }
}
\thanks{$^\ast$The work was done when Yang Jiao was a Research Intern with Meituan. \\
$^\star$Jingjing Chen and Lin Ma are the corresponding authors. {chenjingjing@fudan.edu.cn, forest.linma@gmail.com}}



\begin{abstract}
Referring Image Segmentation (RIS) aims at segmenting the target object from an image referred by one given natural language expression. 
The diverse and flexible expressions as well as complex visual contents in the images raise the RIS model with higher demands for investigating fine-grained matching behaviors between words in expressions and objects presented in images. 
However, such matching behaviors are hard to be learned and captured  when the visual cues of referents (i.e. referred objects) are insufficient, as the referents with weak visual cues tend to be easily confused by cluttered background at boundary or even overwhelmed by salient objects in the image. And the insufficient visual cues issue can not be handled by the cross-modal fusion mechanisms as done in previous work. In this paper, we tackle this problem from a novel perspective of enhancing the visual information for the referents by devising a \textbf{T}wo-stage \textbf{V}isual cues enhancement \textbf{Net}work (TV-Net), where a novel Retrieval and Enrichment Scheme (RES) and an Adaptive Multi-resolution feature Fusion (AMF) module are proposed.
Specifically, RES retrieves the most relevant image from an external data pool with regard to both the visual and textual similarities, and then enriches the visual information of the referent with the retrieved image for better multimodal feature learning.  AMF further enhances the visual detailed information by incorporating the high-resolution feature maps from lower convolution layers of the image.
Through the two-stage enhancement, our proposed TV-Net enjoys better performances in learning fine-grained matching behaviors between the natural language expression and image, especially when the visual information of the referent is inadequate, thus produces better segmentation results.
Extensive experiments are conducted to validate the effectiveness of the proposed method on the RIS task, with our proposed TV-Net surpassing the state-of-the-art approaches on four benchmark datasets. Our code is available at: \url{https://github.com/SxJyJay/TV-Net}.
\end{abstract}





\maketitle

\begin{figure}[t]
    \centering
    \includegraphics[width=\linewidth]{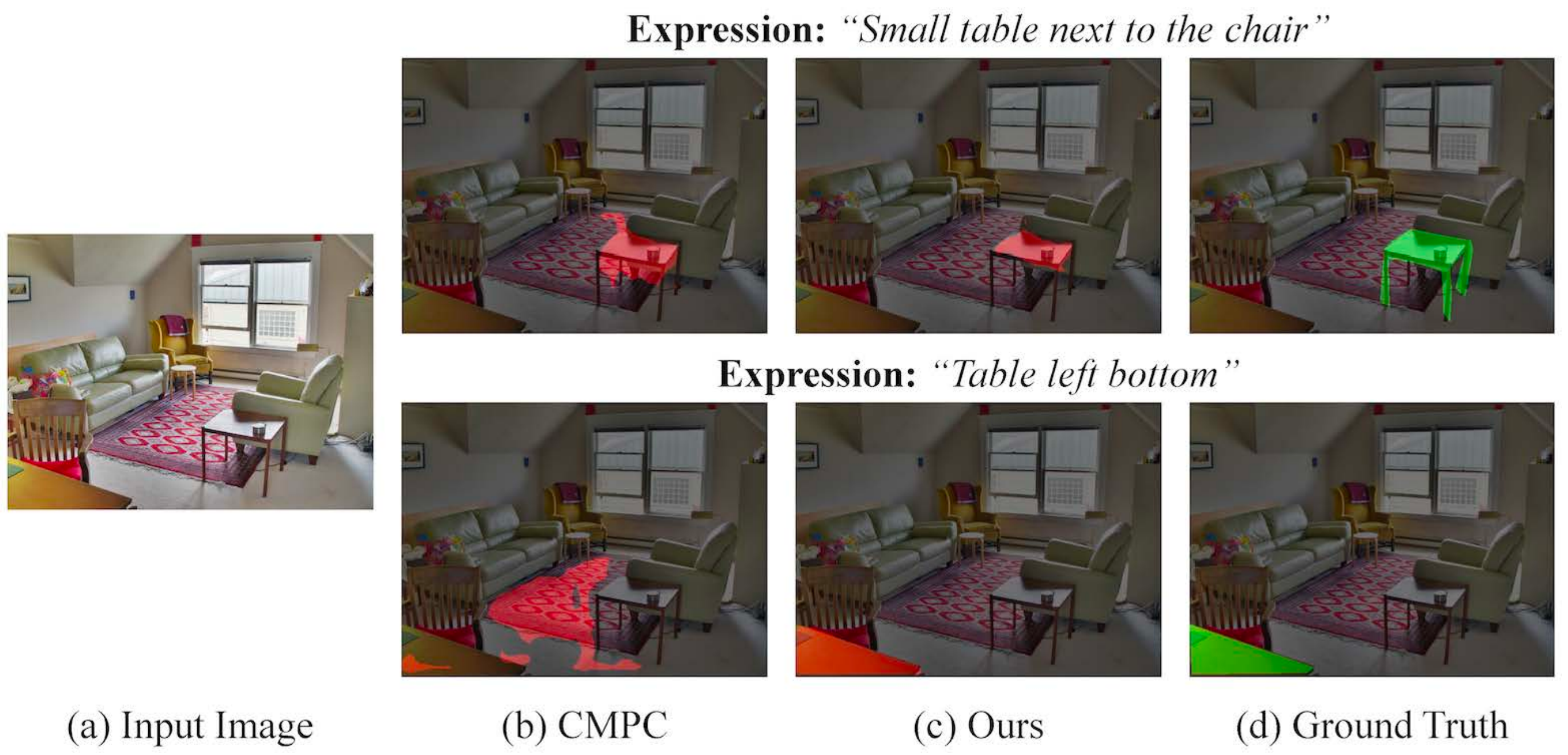}
    \centering
    \caption{Referring image segmentation (RIS) aims to segment the referent (i.e. referred object), which is specified by the given natural language description. Compared with CMPC \cite{huang2020referring}, a state-of-the-art method for RIS, which fails to generate precise or even correct prediction masks for unsalient referents as shown in (b), our proposed TV-Net can produce segmentation masks as shown in (c), which are highly consistent with the ground-truth in (d).}
    \label{Fig1}
\end{figure}

\section{Introduction}
Referring Image Segmentation (RIS)~\cite{DBLP:conf/cvpr/YeR0W19,Chen_2019_ICCV,huang2020referring,DBLP:conf/cvpr/HuFSZL20} has emerged as a prominent multi-discipline research problem. Unlike traditional semantic segmentation that segments the objects of predefined classes, RIS aims to accurately segment the  object specified by one natural language expression. As shown in  Figure~\ref{Fig1}, given two natural language expressions describing two different tables in the same image, RIS aims to  produce the corresponding segmentation masks highlighting the objects that well match the expressions. As such, RIS can facilitate users to freely manipulate segmentation results with various natural language forms, such as color, position, category, which will have a wide application prospect in the fields of interactive image editing \cite{DBLP:conf/mm/ChengGL0G20}, human-robot interaction \cite{DBLP:conf/rss/ShridharH18}, etc.

However, RIS is inherently a challenging problem, which needs  to not only model the characteristics of natural language expression and image but
also exploit the fine-grained interactions between these two modalities. 
To address such a challenge problem, existing approaches spent efforts in devising various cross-modal learning mechanisms, hoping to understand both the textual and visual semantics and further exploit their matching behaviors between words in expressions and visual information in images~\cite{hu2016segmentation,liu2017recurrent,Li_2018_CVPR}. Relying on the attention mechanisms, such matching relationships can be captured by dynamically learning a soft association between the image local patches and words and thereby help produce more accurate segmentation results~\cite{Shi_2018_ECCV,DBLP:conf/cvpr/YeR0W19,Chen_2019_ICCV,DBLP:conf/cvpr/HuFSZL20}.

Despite remarkable progresses have been made with the existing cross-modal  learning mechanisms, 
the segmentation results still remain unsatisfactory,
especially when the visual cues of referents (i.e.,referred objects) are insufficient. As shown in Figure~\ref{Fig1}, given the natural language expression \textit{``small table next to the chair"}, CMPC \cite{huang2020referring}, the state-of-the-art method, is not able to produce a precise segmentation mask, while for \textit{``table left bottom"}, CMPC even fails to correctly locate the specified referent. 
Such unsatisfactory segmentation results are mainly due to two aspects. 
First, the cluttered background visual information, such as the patterned carpet and white floor in the image,  makes the the referents with weak visual cues (e.g., "small table") not distinctive enough, especially at referent boundaries. Second, the most salient object (e.g., "carpet" ) in the image tends to overwhelm the referent of weak visual information (e.g., "table left bottom" in Figure 1(a)), hence yielding a wrong segmentation mask.

To address the aforementioned problems, this paper proposes to enhance the visual information of the referents for RIS, in order to alleviate the ambiguities at the referent boundaries and make the referent not being overwhelmed by the salient objects.
Specifically, we propose a novel \textbf{T}wo-stage \textbf{V}isual cues enhancement \textbf{Net}work (TV-Net), performing  Retrieval and Enrichment Scheme (RES) and Adaptive Multi-resolution feature Fusion (AMF). On the one hand, RES performs the visual cues enhancement by retrieving external relevant images to enrich the visual feature of the referent.
To ensure the strong correlation between the retrieved images and the referent, RES makes use of both the textual and visual features to retrieve the most relevant image for the referred object. 
Afterwards, the enhanced visual features are used for fusing with textual feature to obtain the multimodal features. 
On the other hand, AMF relies on the obtained multimodal feature to incorporate and fuse the meaningful high-resolution visual features in a dynamic and adpative manner, which further complements the visual detailed information of the referent.
Finally, the enhanced features  are utilized to produce the segmentation masks.

We summarize our contributions as follows:
\begin{itemize}
    \item A novel \textbf{T}wo-stage \textbf{V}isual cues enhancement \textbf{Net}work (TV-Net) is proposed for tackling the RIS task. Comparing to the existing works which focus on devising cross-modal learning mechanisms, TV-Net deals with RIS task from one novel perspective of enhancing visual cues of the referents.
    \item TV-Net enhances the referent visual cues through a Retrieval and Enrichment Scheme (RES) and an Adaptive Multi-resolution feature Fusion (AMF) module. Specifically, RES incorporates the visual information semantically related to referents from an retrieved image with the original input image. 
    AMF complements the visual details for referents by  fusing the high-resolution visual features in a dynamic and adpative manner.
    \item Extensive experiments on four benchmark datasets demonstrate that our proposed TV-Net achieves superior performances compared with state-of-the-art methods.
\end{itemize}

\begin{figure*}[t]
\begin{center}
    \includegraphics[width=0.8\linewidth]{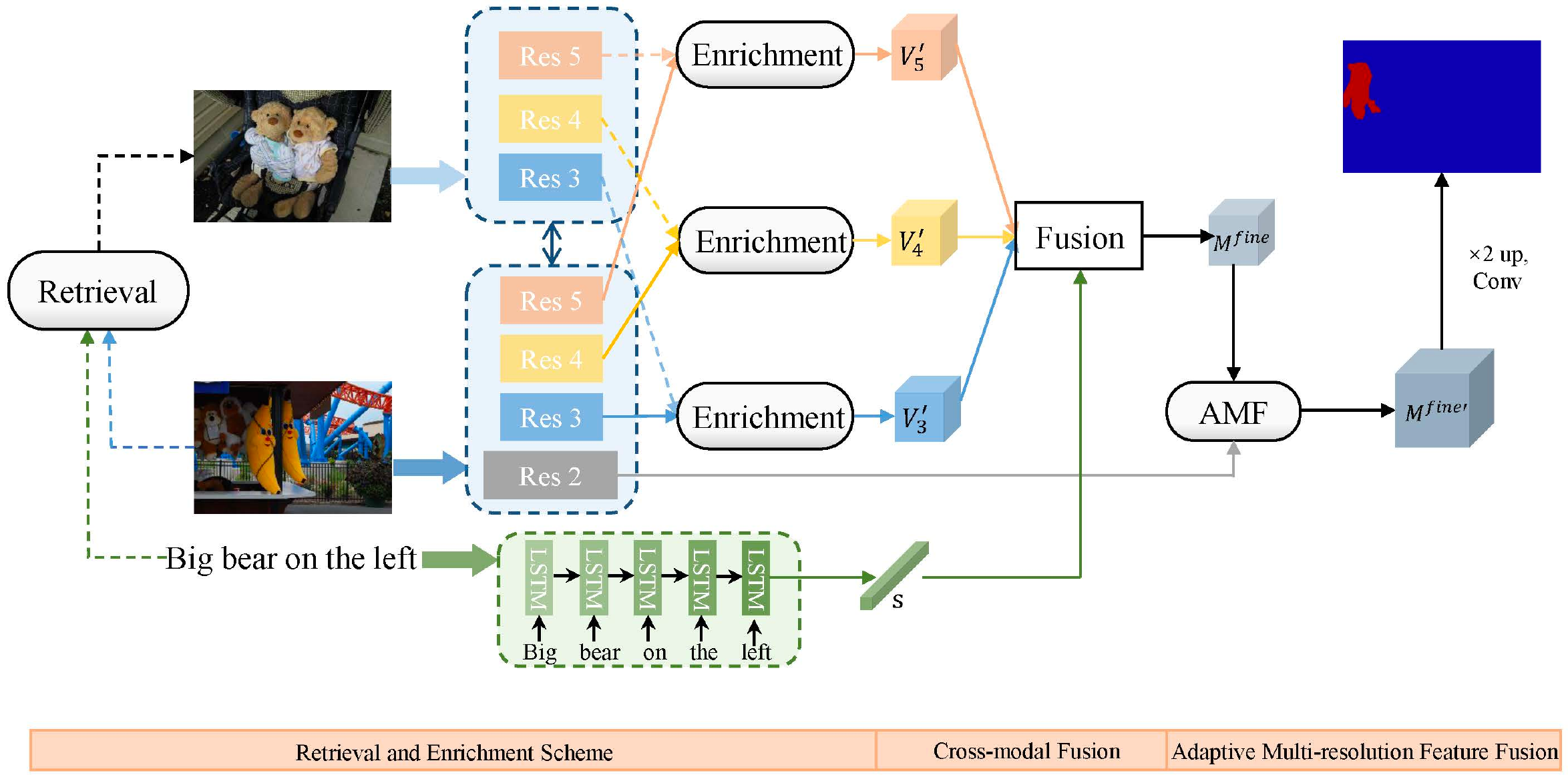}
\end{center}
  \caption{The proposed TV-Net enhances visual cues for referents in two stages. The first stage adopts a novel Retrieval and Enrichment Scheme (RES), where the most relevant image in data pool is retrieved and used for enriching visual information of referents in the input image. Relying on the enriched visual features, cross-modal interaction is performed under the guidance of text. The second stage aims to further enhance the learned multimodal feature by adaptively incorporate visual details from high-resolution visual feature map in Adaptive Multi-resolution feature Fusion (AMF) module. The solid lines represent information flow during end-to-end training process, while the dotted lines indicate information flow in the offline retrieval. The blue dotted part is a siamese-CNN, and notation "$\times$2 up" represents 2 times bilinear interpolation upsampling.}
\label{framework}
\end{figure*}

\section{Related Work}


Referring image segmentation (RIS) aims to generate a segmentation mask for the specific objects referred in a natural language expression. Different from traditional image segmentation, the key for RIS is to establish explicit correspondence between objects in vision modality and words in language modality. 

To address RIS problem, pioneering models~\cite{hu2016segmentation,liu2017recurrent,Li_2018_CVPR} achieve basic multimodal interaction via concatenation-convolution scheme, and devise assortment of structures to further explore the relationships between the two modalities. As the first work to address RIS problem, Hu \textit{et al.}~\cite{hu2016segmentation} directly concatenate and fuse multimodal features from CNN and LSTM. Liu \textit{et al.}~\cite{liu2017recurrent} also follow convolution-operation fashion, and perform the cross-modal interaction in multiple steps to mimic human reasoning process. Similarly, in~\cite{Li_2018_CVPR}, sequentially fusion manner is also adopted to capture semantics in multi-level visual features.

Recently, richer multimodal fusion mechanism has been explored beyond simple concatenation-convolution scheme. Dynamic filters for each word are proposed in~\cite{Margffoy-Tuay_2018_ECCV} to fully exploit the recursive nature of language. Later, attention mechanism is widely used in~\cite{Shi_2018_ECCV,DBLP:conf/cvpr/YeR0W19,Chen_2019_ICCV,DBLP:conf/cvpr/HuFSZL20,huang2020referring,DBLP:conf/eccv/HuiLHLYZH20} as its prevalence in fields of computer vision~\cite{Wang_2018_CVPR,fu2019dual,DBLP:conf/iccv/HuangWHHW019} and natural language processing~\cite{NIPS2017_3f5ee243}. In~\cite{Shi_2018_ECCV}, attention mechanism is utilized to extract key words in language expression and model key-word-aware visual context. Further, Ye \textit{et al.}~\cite{DBLP:conf/cvpr/YeR0W19} concatenate every word with multi-level visual features and devise self-attention mechanism to adaptively focus on informative words in the referring expression and important local patches in the input image. Bi-direction attention module is also investigated for establishing cross-modal correlation in~\cite{DBLP:conf/cvpr/HuFSZL20}. To further explicitly align the vision and language modalities in a co-embedding space, Chen \textit{et al.}~\cite{Chen_2019_ICCV} generate the visual-textual co-embedding map in several recurrent steps. As graph neural network~\cite{DBLP:conf/iclr/VelickovicCCRLB18,ijcai20} presents a new form of mining the relationship between data, Hui \textit{et al.}~\cite{DBLP:conf/eccv/HuiLHLYZH20} and Yang \textit{et al.} introduce graph structure models to achieve efficient message passing in RIS. Moreover, some  works~\cite{huang2020referring,DBLP:conf/eccv/HuiLHLYZH20} also consider the linguistic roles of each word during multimodal interaction process. Words are classified into four categories, and a progressive comprehension process is proposed under the guidance of different type of words in~\cite{huang2020referring}. Hui \textit{et al.} introduces dependency parsing tree~\cite{chen-manning-2014-fast} as prior knowledge to constrain the communication among words, so as to include valid multimodal context and exclude disturbing ones. Different from existing works, we tackle the RIS problem by enhancing visual cues for referents to learn more robust multimodal feature representation.

\section{Method}
The overall architecture of our proposed \textbf{T}wo-stage \textbf{V}isual cues enhancement \textbf{Net}work (TV-Net) is illustrated in Figure~\ref{framework}. Given an image and its referring expression, TV-Net first performs visual cues enhancement through a novel Retrieval and Enrichment Scheme (RES), where the image most relevant to the referred object in the data pool is first retrieved with regard to textual and visual similarity. Then a gated attentional fusion mechanism is utilized to control the auxiliary information flow from the retrieved relevant images.  
Relying on the enriched visual feature from RES and language feature extracted from LSTM, cross-modal fusion is conducted to exploit their relationships and compose them accordingly. Afterwards, an Adaptive Multi-resolution Fusion (AMF) module provides region detail cues in high-resolution visual features complementing to the learned multimodal feature for further enhancement. Finally, the enhanced features after RES and AMF are used for predicting the segmentation masks of referred objects.

\begin{figure}[t]
\begin{center}
    \includegraphics[width=1\linewidth]{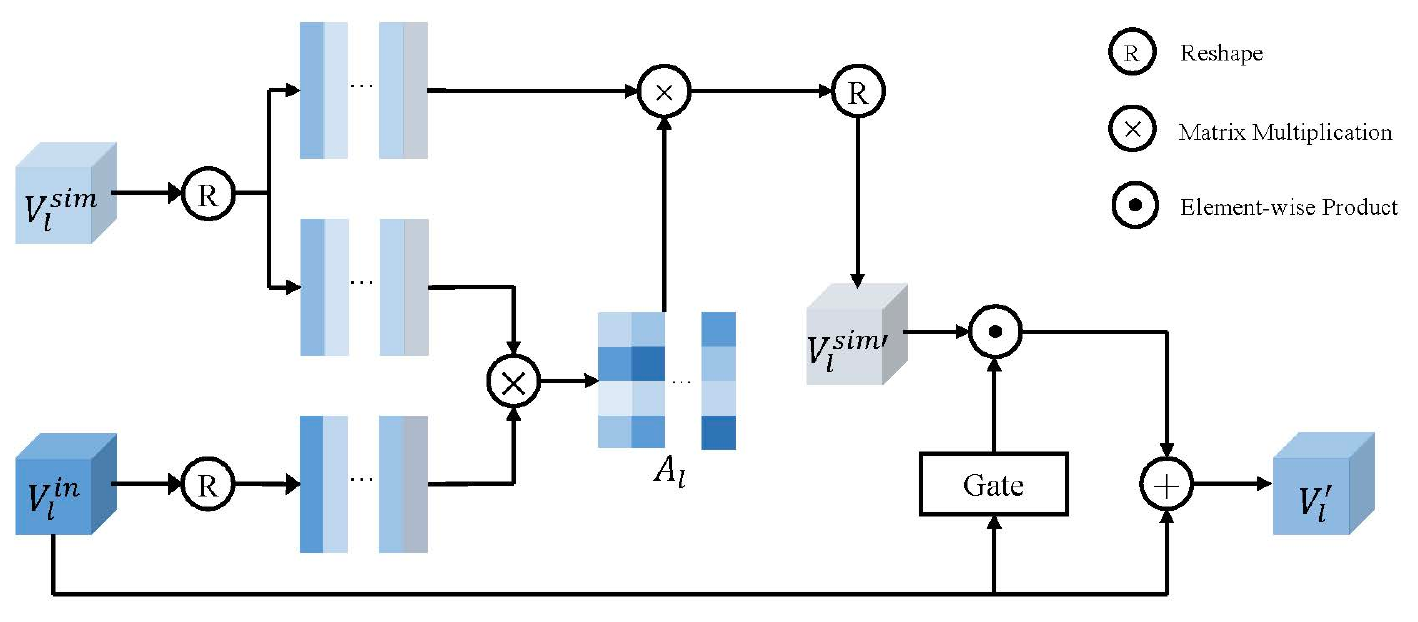}
\end{center}
  \caption{The illustration of detailed process of Enrichment, which adopts a gated attentional fusion mechanism. A cross-image attention is applied to reorganize feature representations at each spatial location of retrieved image according to their correlation with referent. Then, the reorganized visual feature is selectively fused with original visual representation for enhancement.}
\label{Enrich}
\vspace{-10pt}
\end{figure}

\subsection{Retrieval and Enrichment Scheme}
{\bf Retrieval.}\quad For a given pair of input image $I$ and language expression $T$, we retrieve an image $I^{sim}$ semantically relevant to the referred object from data pool for enriching visual information of the input image.
For an accurate retrieval result, both textual and visual similarities between each candidate sample in the data pool and input are measured. Specifically, an image encoder and a text encoder are first used to generate visual and textual features for input and each candidate sample. For description convenience, we regard input as query, and candidate samples in data pool as keys. Accordingly, the extracted visual feature and text feature of input/candidate are called visual query/key and textual query/key.
The visual and textual query pair is denoted as \textit{Q} = \{$\mathbf{q}^{I}$, $\mathbf{q}^{T}$\} and \textit{i}-th visual and textual key pair is denoted as $K_{i}$ = \{$\mathbf{k}_{i}^{I}$, $\mathbf{k}_{i}^{T}$\}, $i$ $\in$ \{1,2,...,N\}, where N is the number of samples in data pool. Next, we match the query \textit{Q} and all N keys $K_{i}$ in the data pool to select the most relevant image $I^{sim}$ for the following referring image segmentation process.  

{\bf Enrichment.}\quad For the input image $I$ together with the retrieved similar image $I^{sim}$, a siamese CNN is adopted to simultaneously extract their visual feature $V^{in}$ and $V^{sim}$.
Here, we define the low-resolution visual feature sets as $\textit{LR}$ = \{$V_{3}$,$V_{4}$,$V_{5}$\} containing feature maps from \textit{res3, res4} and \textit{res5} layers, and high-resolution feature sets as $\textit{HR}$ = \{$V_{1}$,$V_{2}$\} including feature maps from \textit{res1} as well as \textit{res2} layers of the siamese CNN. Semantic information exists in the low-resolution feature maps from deeper layers of CNN while the high-resolution feature maps mainly contain region details, such as edge or texture as implied in~\cite{DBLP:conf/eccv/ZeilerF14}. In order to incorporate visual information semantically relevant to the referred objects, the enrichment process is performed between the low-resolution visual feature sets of input image $\textit{LR}^{in}$ = \{$V_{3}^{in}$,$V_{4}^{in}$,$V_{5}^{in}$\} and the retrieved similar image $\textit{LR}^{sim}$ = \{$V_{3}^{sim}$,$V_{4}^{sim}$,$V_{5}^{sim}$\}.

The details of Enrichment process is depicted in Figure \ref{Enrich}. Given visual feature $V_{l}^{in}$ $\in$ $\mathbb{R}^{h_{l}\times w_{l}\times d_{l}}$ and $V_{l}^{sim}$ $\in$ $\mathbb{R}^{h_{l}\times w_{l}\times d_{l}}$, where $h_{l}$, $w_{l}$ and $d_{l}$ are the height, width and channel dimensions respectively, we reorganize the feature representations at every spatial location in $V_{l}^{sim}$ according to their correlation with referent in $V_{l}^{in}$ via a cross-image attention:
\begin{equation}
    \begin{split}
         &A_{l} = \mathrm{Softmax}((W_{1} \mathbf{v}_{l}^{in})^{\mathrm{T}}(W_{2} \mathbf{v}_{l}^{sim}))\\
         &\mathbf{v}_{l}^{sim'} = A_{l}(W_{3} \mathbf{v}_{l}^{sim})
    \end{split}
\end{equation}
$\mathbf{v}_{l}^{in}$ and $\mathbf{v}_{l}^{sim}$ $\in$ $\mathbb{R}^{h_{l}w_{l} \times d_{l}}$ are the reshaped $V_{l}^{in}$ as well as $V_{l}^{sim}$. $W_{1}$, $W_{2}$ and $W_{3}$ are learnable parameters. The cross-image attention is calculated in the embedded Gaussian version~\cite{Wang_2018_CVPR} to get an attention map $A_{l}$ which measures the importance of every region in $\mathbf{v}_{l}^{sim}$ to $\mathbf{v}_{l}^{in}$. Then $A_{l}$ is used for reorganizing the linear transformed $\mathbf{v}_{l}^{sim}$ to generate visual feature $\mathbf{v}_{l}^{sim'}$ $\in$ $\mathbb{R}^{h_{l}w_{l} \times {d_{l}}}$, which will be served for boosting $\mathbf{v}_{l}^{in}$. Next, to enhance $\mathbf{v}_{l}$ with $\mathbf{v}_{l}^{sim'}$, directly adding or concatenating them might introduce noise, especially in the case of inaccurate retrieved results. Thus, we devise a gate conditioning on the original input visual feature $V_{l}^{in}$ to selectivly incorporate $V_{l}^{sim'}$ into $V_{l}$, which is formulated as bellow, where we omit the reshape transformation from $\mathbf{v}_{l}^{in}$/$\mathbf{v}_{l}^{sim'}$ to $V_{l}^{in}$/$V_{l}^{sim'}$:
\begin{equation}
    V'_{l} = V_{l}^{in}+\sigma(\mathrm{Conv}(V_{l}^{in}))\odot V_{l}^{sim'} \label{sp_tfm}
\end{equation}
$\sigma(\cdot)$ represents sigmoid function. Conv($\cdot$) denotes 1$\times$1 convolution operation, and $\odot$ is the notation of element-wise multiplication.

\subsection{Cross-modal Fusion}
RIS model takes visual and textual features as input, and outputs a predicted segmentation mask. By now the enriched visual features $V'_{l}$ ($l$ $\in$ $\{3,4,5\}$) achieved by RES have been well-prepared. As for language feature, we encode each word in sentence \textit{T} with a LSTM to generate language feature $\mathbf{s}$ $\in$ $\mathbb{R}^{d_{s}}$, where $d_{s}$ is the dimension of language embedding space. Following prior works~\cite{DBLP:conf/cvpr/YeR0W19,huang2020referring,DBLP:conf/cvpr/HuFSZL20}, an 8-D spatial coordinate feature $O \in \mathbb{R}^{h_{l} \times w_{l} \times 8}$ is integrated into the enriched visual features $V'_{l}$ before cross-modal fusion.

As it is commonly assumed that the language refers to high-level visual concepts while leaving low-level visual processing unaffected~\cite{NIPS2017_3f5ee243}, it is natural to aggregate language feature with high-level visual features (i.e., the low-resolution feature maps). Moreover, exploring cross-modal interaction is often formulated as a multi-step progressive process~\cite{DBLP:conf/cvpr/NamHK17,DBLP:conf/cvpr/FanZ18,Mun_2020_CVPR,huang2020referring}. Based on the consideration above, we first adopt Bilinear fusion strategy~\cite{Ben-younes_2017_ICCV} for preliminary multimodal fusion, then refine the raw multimodal feature via integrating features of multiple levels. Such process is formulated as below:
\begin{equation}
    \begin{split}
        &M_{l}^{raw} = (W_{4}\mathbf{s})\odot (W_{5} \otimes \mathrm{Conv}([V'_{l};O]))\\
        &M_{l}^{raw'} = M_{l}^{raw} + \sum_{k \in \{3,4,5\}-\{l\}} \mathrm{Gate}(M_{k}^{raw})
    \end{split}
    \label{CSM_equ}
\end{equation}
where $W_{4}$ $\in$ $\mathbb{R}^{d_{s} \times d_{m}}$ and $W_{5}$ $\in$ $\mathbb{R}^{d_{m} \times d_{m}}$ are weight matrix for linear transformation, $\otimes$ and [;] represents the matrix multiplication and concatenation respectively. Gate($\cdot$) operation has many choices, such as LSTM-type gate in~\cite{DBLP:conf/cvpr/YeR0W19}, vision-guided gate in~\cite{DBLP:conf/cvpr/HuFSZL20} and language-guided gate in~\cite{huang2020referring}. We use the language-guided gate as implemented in ~\cite{huang2020referring}. Finally, we apply ConvLSTM~\cite{NIPS2015_07563a3f} to aggregate multimodal features ($M_{3}^{raw'}$, $M_{4}^{raw'}$ and $M_{5}^{raw'}$) of different levels to obtain a refined multimodal feature $M^{fine}$.

\begin{figure}[t]
\begin{center}
    \includegraphics[width=1\linewidth]{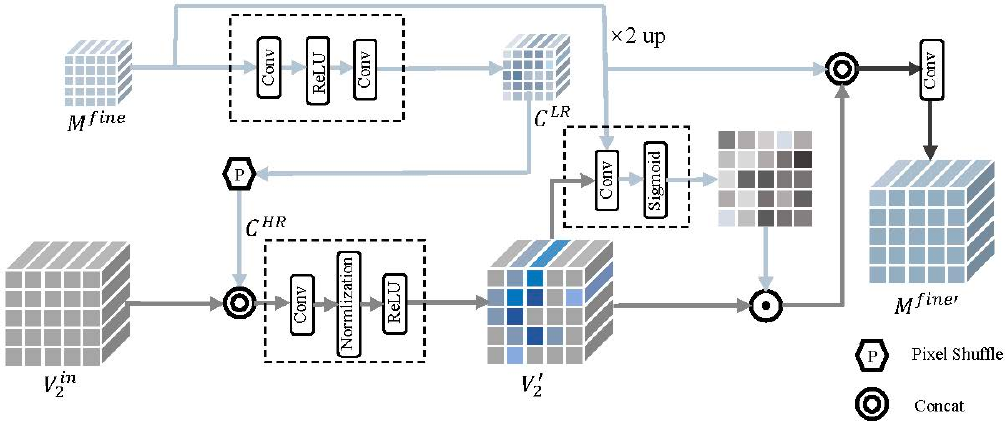}
\end{center}
  \caption{An illustration for proposed AMF module. First, regions irrelevant to the referent in high-resolution visual feature are suppressed by taking in high-level semantics from low-resolution features. And then, a gate consisting of convolution, normalization and ReLU activation operations is devised for reconciling the relative intensities of high-resolution and low-resolution features during their fusion.
  "$\times$2 up" represents 2 times bilinear interpolation upsampling.}
\label{MRE}
\vspace{-16pt}
\end{figure}

\subsection{Adaptive Multi-resolution Feature Fusion}
\label{AMF}
Different from low-resolution visual features, high-resolution visual feature accommodates more region details but are lack of semantics as implied in~\cite{DBLP:conf/eccv/ZeilerF14}. In order to adaptively incorporate region detail cues related to referents in high-resolution visual feature for further enhancement, we investigate a novel multi-resolution feature fusion strategy as illustrated in Figure~\ref{MRE}. Instead of directly fusing the multimodal feature $M^{fine}$ with the high-resolution visual feature $V_{2}^{in}$, we suppress the regions irrelevant to referents in $V_{2}^{in}$ under the guidance of $M^{fine}$, and then dynamically reconcile their relative intensities via a gate during aggregation process. 

Suppose $HR^{in} = \{V_{2}^{in},V_{1}^{in}\}$ represents the high-resolution input image \textit{I}. To reduce computational complexity, we only use $V_{2}^{in}$ for promoting the visual details. 
Specifically, we first encode more visual contents into $C^{LR}$ with convolution and ReLU activation operations, then perform Pixel Shuffle (PS)~\cite{DBLP:journals/corr/ShiCHTABRW16}, a kind of upsampling operation, on $C^{LR}$ to obtain $C^{HR}$ with the same resolution as $V_{2}$.
Afterwards, $C^{HR}$ learns to suppress the regions irrelevant to referents in $V_{2}^{in}$ during training with concatenation and convolution operations:
\begin{equation}
    \begin{split}
        &C^{LR} = \mathrm{Conv}(\mathrm{ReLU}(\mathrm{Conv}(M^{fine}))) \\
        &C^{HR} = \mathrm{PS}(C^{LR}) \\
        &V'_{2} = \mathrm{ReLU}(\mathrm{Norm}(\mathrm{Conv}([C^{HR};V_{2}^{in}])))
    \end{split}
\end{equation}

For fusing the background suppressed feature $V'_{2}$ with upsampled multimodal feature $M^{fine}$, simply adding or concatenating them fails to dynamically and adaptively balance their relative importance at different spatial location. Therefore, we reconcile their intensities using a gate operation, which consists of a sequence of convolution and sigmoid activation operations (i.e., black dotted rectangle parts at the midlle of Figure~\ref{MRE} ):
\begin{equation}
\label{AMF_eq}
    \begin{split}
        &A_{2} = \sigma(\mathrm{Conv}([V'_{2};M^{fine}])) \\
        &M^{fine'} = \mathrm{Conv}([\mathrm{Upsample}(M^{fine});(A_{2} \odot V'_{2})])
    \end{split}
\end{equation}
As shown in Equation (\ref{AMF_eq}), an intensity map $A_{2}$ $\in$ $\mathbb{R}^{h_{2} \times w_{2} \times 1}$ for $V'_{2}$ is first generated with a gate. Each value in $A_{2}$ measures the relative intensity of region detail cues in $V'_{2}$ at current position. Afterwards, the intensity map $A_{2}$ element-wisely multiplies on $V'_{2}$ and concatenate with upsampled $M^{fine}$ to obtain the final enhanced feature $M^{fine'}$, which will be used for the segmentation mask prediction.

\begin{table*}[!ht]
    \caption{Ablation studies of different size of objects on UNC+ val set with Prec@X (X $\in$ \{0.5,0.6,0.7,0.8,0.9\}) and Overall IoU for evaluation.}
    \label{Ablation UNC+ val}
    \begin{center}
    \begin{tabular}{c c c c c c c c c}
        \toprule
        row\_id    &Size    & Method    & Prec@0.5 & Prec@0.6 & Prec@0.7 & Prec@0.8 & Prec@0.9 & Overall IoU\\
        \midrule
        1    &\multirow{5}{*}{All}    & CMPC~\cite{huang2020referring}      & 58.47   & 52.28    & 43.43    & 28.94     & 7.24  & 49.56 \\
        2    &~       & Baseline                  & 55.19   & 48.02    & 38.84    & 25.20     & 6.25   & 48.92\\
        3    &~       & Baseline+RES              & 55.70   & 48.51    & 39.42    & 25.14     & 6.14   & 48.89\\
        4    &~       & Baseline+RES+Cct          & 57.69   & 51.17    & 42.61    & 29.29     & 8.20   & 49.93\\
        5    &~       & Baseline+RES+AMF          &\textbf{58.62}   &\textbf{52.71}    &\textbf{44.26}    &\textbf{31.89}     &\textbf{10.95} &\textbf{50.30}\\
        \midrule
        6    &\multirow{5}{*}{Small}  & CMPC~\cite{huang2020referring}      & 44.79   & 38.51   & 29.22     & 14.59     & 1.29 & 30.96 \\
        7    &~       & Baseline                  & 39.97   & 31.96   & 22.85     & 11.62     & 0.82   & 29.63\\ 
        8    &~       & Baseline+RES              & 41.09   & 32.83   & 23.45     & 11.19     & 0.73   & 29.86\\
        9    &~       & Baseline+RES+Cct          & 41.74   & 34.04   & 26.03     & 15.15     & 1.51   & 29.03\\
        10   &~       & Baseline+RES+AMF          &\textbf{46.47}   &\textbf{39.50}   &\textbf{30.77}     &\textbf{19.45}     &\textbf{3.27}  &\textbf{31.54}\\ 
        \bottomrule
    \end{tabular}
    \end{center}
\end{table*}

\section{Experiments}

\subsection{Experimental Setup}
{\bf Datasets.}\quad We conduct extensive experiments on four benchmark datasets for RIS: UNC~\cite{unc_refexp}, UNC+~\cite{unc_refexp}, G-Ref~\cite{google_refexp} and ReferIt~\cite{KazemzadehOrdonezMattenBergEMNLP14}. 
Images in UNC, UNC+ and G-Ref are all collected from MS COCO~\cite{lin2014microsoft}, while ReferIt selects images from IAPR TC-12~\cite{journals/cviu/EscalanteHGLMMSPG10}. UNC contains 19,994 images with 142,209 language expressions for 50,000 segmentation regions. Objects of the same category may appear more than once in each image. UNC+ consists of 141,564 descriptions for 49,856 objects in 19,992 images. No location words appear in UNC+ language descriptions, which means the category and attributes are only hints for segmenting referred objects. G-Ref comprises of 104,560 expressions referring to 54,822 objects in 26,711 images. The length of description sentence is longer as well as diversiform. ReferIt includes 19,894 images with 96,654 objects referred by 130,525 language expressions, and Referents in it can be objects or stuff (e.g., ground and sky).

\begin{figure*}[!t]
\begin{center}
    \includegraphics[width=0.9\linewidth]{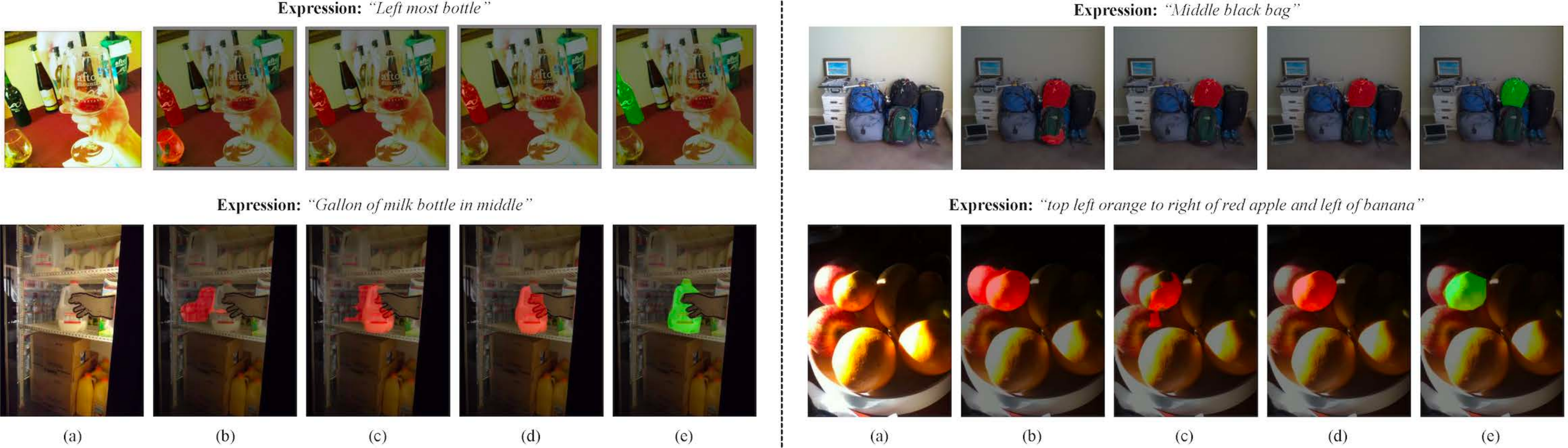}
\end{center}
  \caption{Prediction results by our proposed module.(a) Original image. (b) Results predicted by our baseline. (c) Results predicted by Baseline+RES. (d) Results predicted by Baseline+RES+AMF (Our full model). (e) Ground-Truth.}
\label{Qualitative Results Abl}
\end{figure*}

{\bf Implementation details.}\quad Our network consists of two execution stages. The first stage aims to retrieve a similar image offline and the second stage is an end-to-end training process. Details of experiment settings are listed as follows:
In retrieval process, we adopt VGG16~\cite{vgg} and DistilBERT~\cite{DBLP:journals/corr/abs-1910-01108} with weights pretrained on ImageNet~\cite{DBLP:conf/cvpr/DengDSLL009} and BookCorpus~\cite{moviebook} to preprocess the image and text. We simply use cosine similarity as measurement. Note that the matching measurement is not restricted to cosine similarity, any other strategies or well-designed models for calculating the similarity can serve as substitution to obtain more precise retrieval results in the future. Specifically, we first reduce the searching space by matching all text keys $\mathbf{k}_{i}^{T}$ in data pool with text query $\mathbf{q}^{T}$ to select 20 most 
relevant candidate samples. Further, the most resemble image $I^{sim}$ among these 20 samples is picked according to visual similarity. The whole retrieval process is finished offline, which means that the retrieved similar image for every paired input image and referring expression is prepared for the subsequent end-to-end training. 

During the end-to-end training, every input image is resized and zero-padded to 320$\times$320, and fed into DeepLab ResNet-101v2~\cite{7913730} pretrained on PASCAL-VOC dataset~\cite{DBLP:journals/ijcv/EveringhamGWWZ10} following the previous work~\cite{DBLP:conf/cvpr/YeR0W19,huang2020referring,Chen_2019_ICCV}. We extract feature maps from DeepLab ResNet-101v2 layers \textit{res3, res4, res5} as low-resolution feature set and \textit{res1, res2} as high-resolution feature set. The spatial scale of each feature map is 40$\times$40 in low-resolution feature set, and 80$\times$80 in high-resolution feature set. As for language processing, GloVe~\cite{pennington-etal-2014-glove} word embeddings pretrained on Common Crawl 840B tokens are firstly used for constructing word embedding look-up table. Then a LSTM takes GloVe word embeddings as input for capturing context information. Following previous work~\cite{DBLP:conf/cvpr/YeR0W19,hu2016segmentation}, We set the dimension of hidden state in each LSTM cell as 1000. Adam~\cite{DBLP:journals/corr/KingmaB14} optimizer is initialized with learning rate of 0.00025 and equipped with polynomial decay with power of 0.9. And the weight decay is set as 0.0005. Considering the GPU memory limits, we set batch size as 1. Ordinary cross-entropy loss is adopted to supervise training process. It is worth noting that we don't use DenseCRF~\cite{NIPS2011_beda24c1} for further refinement during inference like previous work \cite{DBLP:conf/cvpr/YeR0W19,DBLP:conf/cvpr/HuFSZL20,huang2020referring}, but still outperforms their models with DenseCRF for prediction mask refinement.

{\bf Evaluation metrics.}
\label{metric}
\quad Following previous work~\cite{DBLP:conf/cvpr/YeR0W19,Chen_2019_ICCV,DBLP:conf/cvpr/HuFSZL20,huang2020referring}, we also use two typical metrics for evaluation: Overall Intersection-over-Union (Overall IoU) and Prec@X, where X $\in$ \{0.5,0.6,0.7,0.8,0.9\}. However, previous works don't discuss the intrinsic characteristics of these 2 metrics. Here we first state the formulas for Overall IoU and Prec@X, then analyze the specific model capability that each of the metric measures.

The formulas of Overall IoU and Prec@X are shown as below:
\begin{align}
    &\mathrm{Overall\_IoU} = \frac{\sum_{i=1}^{M}\mathrm{I}_{i}}{\sum_{i=1}^{M}\mathrm{U}_{i}} \label{Overall_IoU}\\
    &\mathrm{Prec@X} = \frac{\sum_{i=1}^{M}\mathbbm{1}\{\mathrm{IoU}_{i} \geq X\}}{M} \label{PrecX}
\end{align}
where $\mathrm{I}_{i}$, $\mathrm{U}_{i}$ and $\mathrm{IoU}_{i}$ represent the intersection region, union region and intersection region over union region of \textit{i}-th data in test set. $M$ is the size of test set. As implied in Equation (\ref{Overall_IoU}), Overall IoU calculates the total intersection region over union region of all samples. It mainly reflects model performance of pixel-level, which does harm to objects of small size, because small objects with limited pixels are easily dominated by the large objects with amounts of pixels.
On the contrary, Prec@X mainly measures the model performance of object level, for it clearly calculates the percentage of objects in test set whose IoU higher than the threshold X.

\subsection{Ablation Studies}
\label{Abl}
The proposed network achieves visual boost by RES and AMF. To comprehend the effectiveness of each of them, we conduct ablation experiments on UNC+ and G-Ref dataset via quantitative as well as qualitative comparison. Due to page limitation, only ablation experiment results on UNC+ val set are shown in this paper with Table~\ref{Ablation UNC+ val}, and the remaining results can refer to the supplementary material. Moreover, in order to especially investigate the effectiveness on the referents weak in visual cues, we also test on small objects, a special case of objects with weak visual cues, inside each dataset. Here, we roughly define the concept of \textit{"small"} as: \textit{objects with mask accounting for less than 5\% area of the whole image}. For the Prec@X measure model performance of object level as stated in Section~\ref{metric}, we make analysis mainly based on Prec@X in Table~\ref{Ablation UNC+ val}.

{\bf RES.}
\label{ab_RES}
\quad We first explore the effectiveness of proposed RES on UNC+ val set, and the results are shown in rows 2, 3 of Table~\ref{Ablation UNC+ val}. Our baseline adopts DeepLab ResNet-101v2 and LSTM as backbone, and only language feature and 3 low-resolution visual features from \textit{res3}, \textit{res4}, \textit{res5} layers of ResNet-101v2 are used for cross-modal fusion and final prediction. For fair of comparison, we only add RES while keep other settings the same as baseline, which is denoted as Baseline+RES in Table~\ref{Ablation UNC+ val}. By comparing row 2 and row 3, we found that RES brings improvement on Prec@0.5, Prec@0.6 and Prec@0.7, but decreases Prec@0.8 and Prec@0.9. Such results is reasonable for RES introduces external relevant visual information that makes referents more indicative, thus achieves better performance when threshold of IoU is relatively low. Nevertheless, these introduced visual information may not be distributed to referents in original image consistently at spatial domain, which hinders the model from generating a precise mask. The Prec@0.8 and Prec@0.9, who set a high threshold, only accept generated mask highly consistent with ground-truth as correct prediction, therefore these 2 metrics drop after adding RES.

\begin{table*}[!ht]
    \caption{Comparison with state-of-the-art (SOTA) methods on four benchmark datasets in terms of Overall IoU. "+DCRF" means utilizing DenseCRF for refinement. Our model don't use DenseCRF but still outperforms most of SOTA methods.}
    \label{ordinary results}
    \begin{center}
    \begin{tabular}{c c c c c c c c c}
        \toprule
        Method                              & \quad & UNC   & \quad & \quad & UNC+  & \quad & G-Ref & ReferIt\\
        \quad                               & val   & testA & testB & val   & testA & testB & val   & test\\ 
        \midrule
        LSTM-CNN~\cite{hu2016segmentation}  & -     & -     & -     & -     & -     & -     & 28.14 & 48.03\\
        RMI+DCRF~\cite{liu2017recurrent}    & 45.18 & 45.69 & 45.57 & 29.86 & 30.48 & 29.50 & 34.52 & 58.73\\
        DMN~\cite{Margffoy-Tuay_2018_ECCV}  & 49.78 & 54.83 & 45.13 & 38.88 & 44.22 & 32.29 & 36.76 & 52.81\\
        KWA~\cite{Shi_2018_ECCV}            & -     & -     & -     & -     & -     & -     & 36.92 & 59.09\\
        RRN+DCRF~\cite{Li_2018_CVPR}        & 55.33 & 57.26 & 53.95 & 39.75 & 42.15 & 36.11 & 36.45 & 63.63\\
        MAttNet~\cite{Yu_2018_CVPR}         & 56.51 & 62.37 & 51.70 & 46.67 & 52.39 & 40.08 & -     & -\\
        CMSA+DCRF~\cite{DBLP:conf/cvpr/YeR0W19}        & 58.32 & 60.61 & 55.09 & 43.76 & 47.60 & 37.89 & 39.98 & 63.80\\
        STEP~\cite{Chen_2019_ICCV}          & 60.04 & 63.46 & 57.97 & 48.19 & 52.33 & 40.41 & 46.40 & 64.13\\
        BRINet~\cite{DBLP:conf/cvpr/HuFSZL20}          & 60.98 & 62.99 & 59.21 & 48.17 & 52.32 & 42.11 & 47.57 & 63.11\\
        BRINet+DCRF~\cite{DBLP:conf/cvpr/HuFSZL20}     & 61.35 & 63.37 & 59.57 & 48.57 & 52.87 & 42.13 & 48.04 & 63.46\\
        CMPC+DCRF~\cite{huang2020referring} & 61.36 & 64.53 & 59.64 & 49.56 & 53.44 & 43.23 & 49.05 & 65.53\\
        LSCM+DCRF~\cite{DBLP:conf/eccv/HuiLHLYZH20}&61.47&64.99& 59.55&49.34& 53.12 & 43.50 & 48.05 &\textbf{66.57}\\
        \midrule
        Ours                                 &\textbf{61.87} &\textbf{65.61} &\textbf{60.10} &\textbf{50.30} &\textbf{54.43} &\textbf{43.52} &\textbf{49.92} & 65.38\\
        \bottomrule
    \end{tabular}
    \end{center}
\end{table*}

\begin{table}[!ht]
    \caption{Comparison with SOTA methods on small referents in each of the four benchmark datasets using Overall IoU as metric. Our model improves the performance on small referents significantly.}
    \label{small results}
    \begin{center}
    \begin{tabular}{c c c c c}
        \toprule
        Dataset     & Setname       & \quad     & Methods   & \quad \\
        \quad       & \quad         & CMPC~\cite{huang2020referring} & LSCM~\cite{DBLP:conf/eccv/HuiLHLYZH20} & Ours \\
        \midrule
        \quad       & val           & 43.40    & 42.47    &\textbf{44.69} \\
        UNC         & testA         & 43.56    & 44.55    &\textbf{46.02} \\ 
        \quad       & testB         & 38.09    & 37.65    &\textbf{39.53} \\
        \midrule
        \quad       & val           & 30.96    & 29.30    &\textbf{31.54} \\
        UNC+        & testA         & 34.90    & 33.93    &\textbf{35.59} \\
        \quad       & testB         & 22.97    & 23.74    &\textbf{23.92} \\ 
        \midrule
        G-Ref       & val           & 29.67    & 29.83    &\textbf{30.35} \\
        \midrule
        ReferIt     & test          & 29.75    &\textbf{31.14}    &30.77  \\
        \bottomrule
    \end{tabular}
    \end{center}
    \vspace{-8pt}
\end{table}

We further test the effectiveness of RES on small referents in rows 7, 8 of Table~\ref{Ablation UNC+ val}. By comparing the relative gain on all sizes of referents (row 2 and row 3) and small referents (row 7 and row 8), we notice that improvements brought by RES on small objects (1.12\% for Prec@0.5, 0.87\% for Prec@0.6, 0.60\% for Prec@0.7) are more obvious than all sizes of objects (0.51\% for Prec@0.5, 0.49\% for Prec@0.6, 0.58\% for Prec@0.7), which indicates the external knowledge is more helpful for referents deficient in visual cues (i.e. small objects in our experiment).

{\bf AMF.}
\quad In rows 3, 4, 5 of Table \ref{Ablation UNC+ val}, we carefully control variable to verify the effectiveness of AMF. As described in Sec\ref{ab_RES}, our baseline equipped with RES (Baseline+RES) only takes visual features from \textit{res3}, \textit{res4} and \textit{res5} as input, therefore, it is unfair to directly compare our final model (Baseline+RES+AMF) that also uses \textit{res2} feature with Baseline+RES model. Based on the above consideration, we add another experiment that directly concatenate the high-resolution visual feature map from \textit{res2} with learned low-resolution multimodal feature, and use concatenation followed by a 1$\times$1 convolution layer for fusion, which is denoted as "Baseline+RES+Cct" in Table \ref{Ablation UNC+ val}. As illustrated in row 4 of Table~\ref{Ablation UNC+ val}, although directly concatenate high-resolution feature can achieve considerable improvement, our proposed AMF module even outperforms the Baseline+RES+Cct with a large margin. Moreover, the improvements on Prec@0.8 and Prec@0.9 are particularly evident, which implies that AMF more focus on generating a precise mask for referents.

Likewise, we conduct experiments on small objects (row 8, 9, 10 in Table~\ref{Ablation UNC+ val}). As shown in Table~\ref{Ablation UNC+ val}, our AMF improves all metrics for small referents more obviously than all sizes of objects, and the Prec@0.9 for small objects even 2 times higher than the Baseline+RES+Cct.

{\bf Qualitative Results. }
\quad To intuitively understand the behavior of RES and MRE, we visualize the results predicted by different version of our model (Baseline, Baseline+RES, Full model) in Figure \ref{Qualitative Results Abl}.  As shown in the left part of the second row in Figure\ref{Qualitative Results Abl}, our baseline model pays more attention to the mineral water bottles near the milk bottle. We consider the reason for such results might be that large part of milk bottle is occluded by the hand, thus model unable to recognize it and choose the mineral water bottle in the middle as the referred object. After RES enriches visual information of referent, model can locate the correct referred object as shown in the "milk bottle" case of Figure \ref{Qualitative Results Abl}(c). Our AMF further refines the prediction results and generates a much better prediction mask than our baseline. The other examples in Figure \ref{Qualitative Results Abl} show a similar phenomenon, the difference is that RES helps model to select the correct object among multiple objects of the same category.

\begin{figure*}[t]
\begin{center}
    \includegraphics[width=0.9\linewidth]{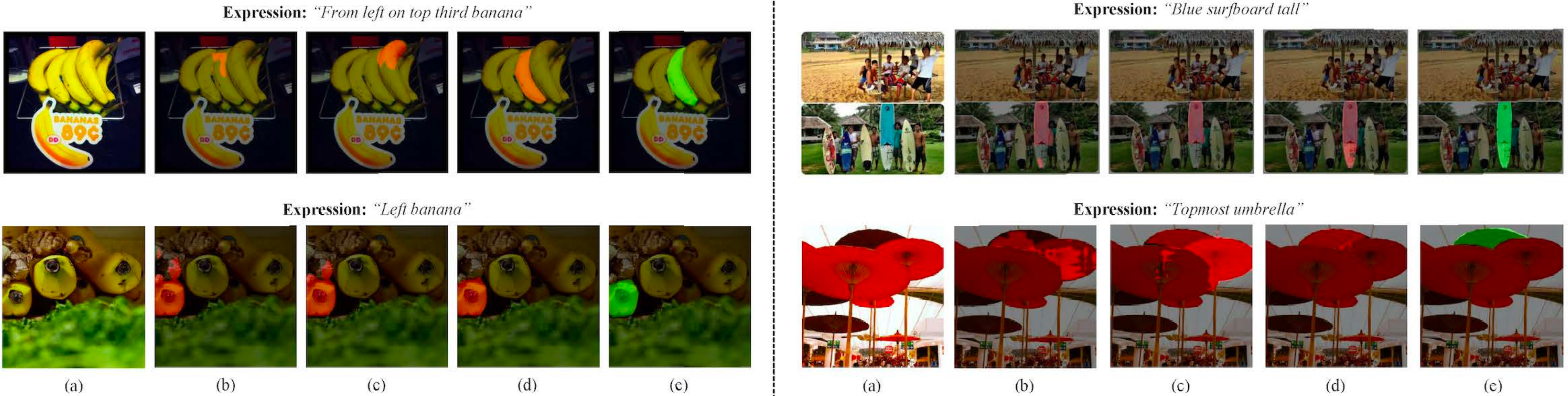}
\end{center}
  \caption{Comparison of our results and two state-of-art methods (LSCM and CMPC).(a) Original image. (b) Results predicted by LSCM. (c) Results predicted by CMPC. (d) Our prediction results. (e) Ground-Truth.}
\label{Qualitative Results SOTA}
\end{figure*}

\begin{figure}[t]
\begin{center}
    \includegraphics[width=0.9\linewidth]{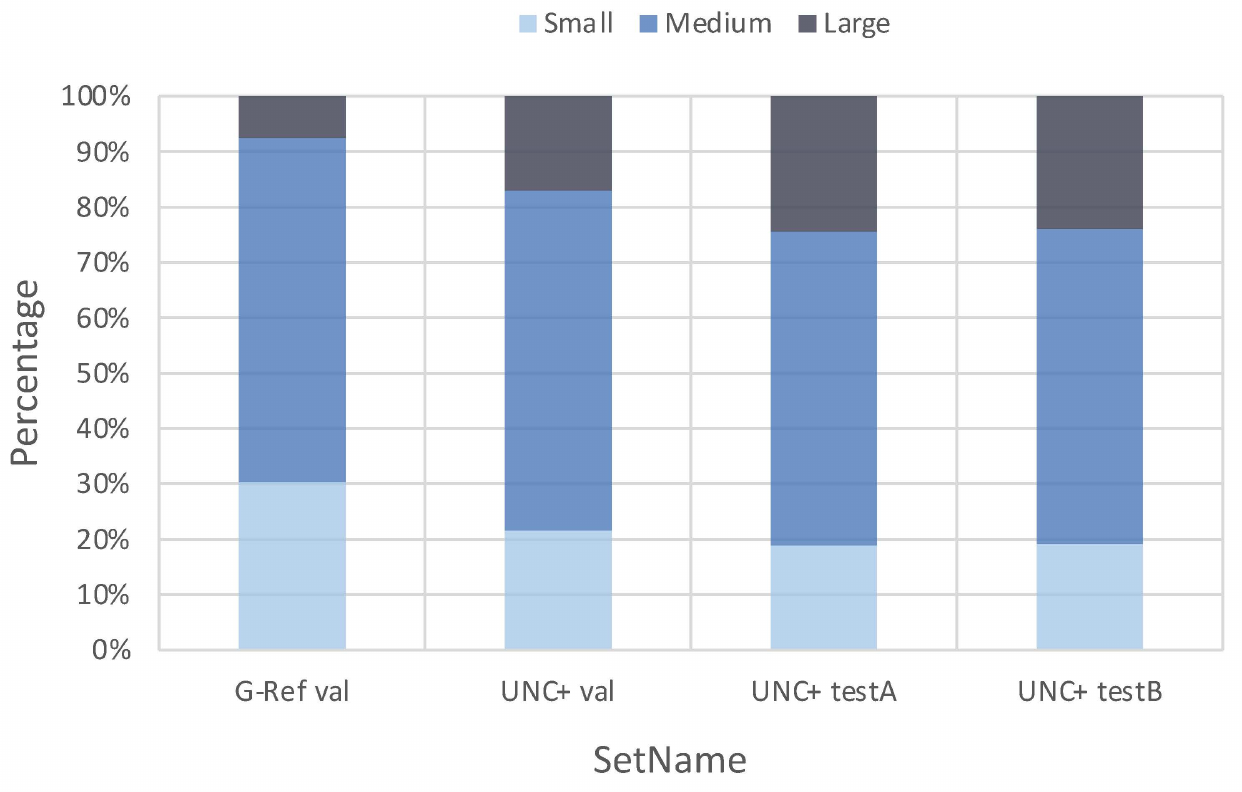}
\end{center}
  \caption{The distribution of objects size in G-Ref val and UNC+ val, testA, testB sets. The proportion of small objects in the Gref val set is higher than that in UNC+ val, testA and testB sets. The definition for \textit{"small"} is the same as in Section~\ref{Abl}, and \textit{"medium"} is defined as object whose mask takes 5\% to 10\% area of the image. The rest of cases is regarded as \textit{"big"} objects.  
  }
\label{Distribution}
\vspace{-14pt}
\end{figure}

\vspace{-6pt}
\subsection{Comparisons with State-of-the-art Approaches}
We compare the proposed approach with several State-Of-The-Art(SOTA) method \cite{DBLP:conf/cvpr/YeR0W19,huang2020referring,DBLP:conf/cvpr/HuFSZL20}, including 2 strongest models (i.e. CMPC~\cite{huang2020referring} and LSCM~\cite{DBLP:conf/eccv/HuiLHLYZH20}). The comparison results are shown in Table~\ref{ordinary results}.
From the result, We observe that our pure model even achieve better results than SOTA methods using DenseCRF~\cite{NIPS2011_beda24c1} for further refinement on most benchmark datasets. Remarkably, particularly obvious gain is obtained on G-Ref by our model, while the performance improvements are not as high on UNC and UNC+. One reason is that the UNC is easier as indicated by the metrics (Nearly 16\% higher Overall IoU than G-Ref), and relatively early work (e.g. CSMA~\cite{DBLP:conf/cvpr/YeR0W19}, STEP~\cite{Chen_2019_ICCV}) already perform well. Another reason is that small objects appear more frequently in G-Ref than in UNC+ as shown in~\ref{Distribution}, therefore, the G-Ref is better able to benefit from our model. 
In addition, two methods (STEP~\cite{Chen_2019_ICCV} and LSCM~\cite{DBLP:conf/eccv/HuiLHLYZH20}) adopting high-resolution visual features should be especially noticed. STEP~\cite{Chen_2019_ICCV} iteratively fuses 5 levels of visual features for 25 times and LSCM~\cite{DBLP:conf/eccv/HuiLHLYZH20} sequentially aggregates 4 levels of visual features through bottom-to-up and top-to-down style as in~\cite{Liu_2018_CVPR}. Instead of repeatedly utilizing high-resolution visual feature, we directly feed high-resolution into proposed AMF block, but still yields 3.68\% and 2.03\% overall IoU boost against STEP and LSCM+DCRF on G-Ref val set, which indicates the effectiveness of our design.

To further investigate the effectiveness of proposed TV-Net on the cases weak in visual cues, we compare our model with two strong SOTA models (CMPC and LSCM) on small referents within each dataset in Table~\ref{small results}. Besides, in order to evaluate pure model ability, we don't use DenseCRF~\cite{NIPS2011_beda24c1} as further refinement in LSCM as well as CMPC. As illustrated in Table \ref{small results}, our method surpasses both CMPC and LSCM with a large margin on most of the benchmarks. In general, our model achieves significant superior results on cases weak in visual cues while also gets better performance on all objects than SOTA methods.

The qualitative results of LSCM, CMPC and our model are depicted in Figure\ref{Qualitative Results SOTA}. From the first example in the left part of row one, we observe that CMPC unables to find the correct referred banana. Compared with CMPC, although LSCM can correctly locate target banana, it generates a prediction mask far from complete. Our prediction result is highly consistent with the ground-truth.
In the example right part of row one, LSCM and CMPC ignore the lower part of the surfboard, while our model can accurately segment the whole surfboard, which proves that our model can well handle objects deficient in visual cues in RIS.  

\begin{figure}[!t]
\begin{center}
    \includegraphics[width=0.9\linewidth]{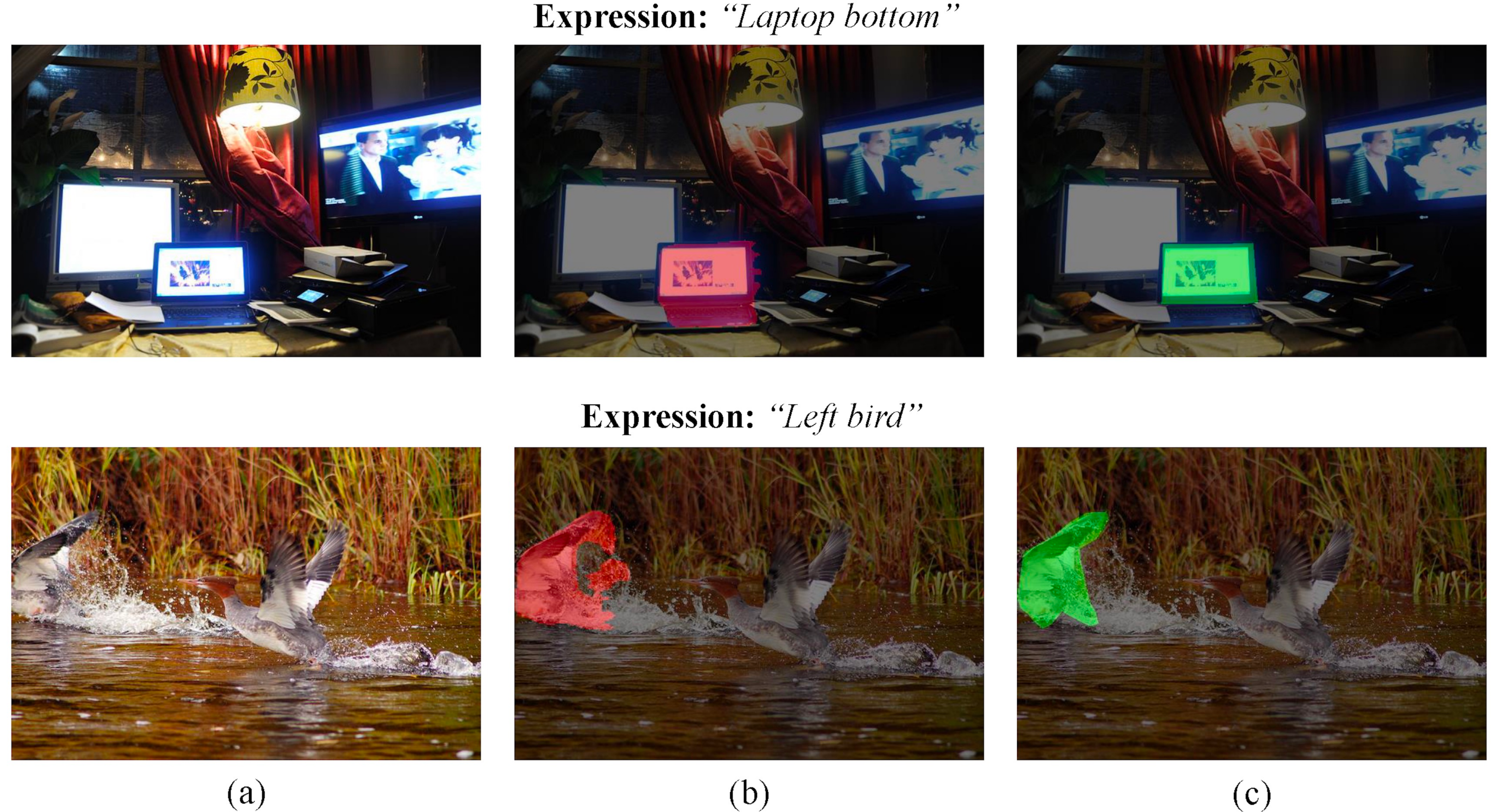}
\end{center}
  \caption{Some failure cases of our model. (a) Original image. (b) Results predicted by our model. (c)  Ground-Truth. These failures are either caused by inaccurate language expression (example in first row), or the lack of visual feature and boundary ambiguity (example in second row).}
\label{Failure Cases}
\vspace{-16pt}
\end{figure}

\subsection{Failure Cases}
We also demonstrate some interesting failure cases in Figure~\ref{Failure Cases}. We argue that such failure cases are mainly due to following two aspects. First, the inaccurate language expressions mislead segmentation prediction. For example, the ground-truth mask for expression \textit{"laptop bottom"} only includes the screen part of referred laptop, while our model segments the complete laptop. Second, extreme ambiguity at boundaries (e.g., splash of water around bird wings in the second row of Figure~\ref{Failure Cases}) deteriorates segmentation results.

\section{Conclusion}
In this paper, we dedicate on the insufficient visual cues problem and propose a \textbf{T}wo-stage \textbf{V}isual cues enhancement \textbf{Net}work (TV-Net) that first utilizes external data to enrich visual information through a novel Retrieval and Enrichment Scheme (RES), and further enhances visual details for referred objects by dynamically incorporating region detail cues in high-resolution visual features with an Adaptive Multi-resolution feature Fusion (AMF) module. Our model achieves improvements on four benchamrk datasets compared with the state-of-the-art models. And such improvements are more evident in small objects, a special case of objects with weak visual cues, which thus demonstrates effectiveness of our model.

\section{Acknowledgement}
This work was supported by National Natural Science Foundation of Project (62072116) and Shanghai Pujiang Program (20PJ1401900).

\bibliographystyle{ACM-Reference-Format}
\bibliography{TV-Net}

\end{document}